# Visual Attention for Behavioral Cloning in Autonomous Driving

Sourav Pal*, Tharun Mohandoss *, Pabitra Mitra
IIT Kharagpur, India

## ABSTRACT

The goal of our work is to use visual attention to enhance autonomous driving performance. We present two methods of predicting visual attention maps. The first method is a supervised learning approach in which we collect eye-gaze data for the task of driving and use this to train a model for predicting the attention map. The second method is a novel unsupervised approach where we train a model to learn to predict attention as it learns to drive a car. Finally, we present a comparative study of our results and show that the supervised approach for predicting attention when incorporated performs better than other approaches.

## 1. INTRODUCTION

Autonomous vehicles are capable of sensing their surrounding environment and navigating from one place to another without significant human intervention. There has been a lot of research on autonomous driving both in the past as well as in recent years. Pomerleau et al. [1] in the seminal project named ALVINN first demonstrated the possibility of using neural networks to train autonomous vehicles. The recent advancements can be very well attributed to progress in deep learning techniques which has been boosted by the abundance of large amounts of data and the computational power that have become available. However, real time computation in presence of large volumes of data remains a challenge.

In case of the autonomous driving scenario, the autonomous agent is overwhelmed with a huge amount of information from cameras, which act as sources of input for sensing the environment around the vehicle. There are several challenges, that come into play including that of large computation expenses and noisy data from the sensors. The scene in front of a vehicle contains vivid information about several objects, which makes the scene intractable.

To this extent we propose that the incorporation of visual saliency as an input to the autonomous agent can improve its performance. The motivation for this stems from the fact that when human drivers navigate a vehicle through a road, they do not pay equal visual attention to everything that is present in their field of sight. As an example, let us consider that there are two objects namely a traffic signal and a building present in the field of sight in front of a human driver, then the driver is most likely to pay more attention to the traffic signal rather than the building which is present out of the road. We model this attention in terms of saliency (visual attention) of the image of the road in front of the autonomous agent.

One of the earliest works on visual attention modeling was done by Itti et al. [2] and was based on the feature integration theory by Treisman et al. [3]. Recent improvements in saliency or attention prediction has been achieved through the use of convolution neural networks (CNNs) [4]. Recent models like SalNet [5], SALGAN [6], DeepFix [7], DeepGaze [8], ML-net [9] and Salicon [10] use a bottom up approach to predicting saliency by learning hierarchies of visual features directly from pixel level data.Saliency in natural images has been widely studied and recent works like that of SALGAN [6] and SalNet [5] show that they perform really well when compared to the ground truth. Saliency of road scenes on real road datasets has also been studied in the recent past such as in [11] and [12]. We consider two approaches for predicting saliency in the task-based approach where the task is to drive a vehicle safely from one point to another while optimizing time and other rewards that lead to human satisfaction.

In the first supervised learning method, we let human drivers drive a car in the simulator Deepdrive [13] and record the positions of their eye-gaze on the screen (effectively the image of the road in front of the car) as they drive. We used Tobi Pro[1], a commodity hardware and open source software OGAMA(OpenGazeAndMouseAnalyzer) [14] to record the eye gaze data. We then use this data as a gold standard to train a CNN to predict the pixel-wise saliency/attention for driving images.

*These two authors contributed equally.

In the second method, we have used an unsupervised approach where the model learns attention/saliency models as it learns to drive. The motivation for this approach is that we humans are not provided any supervised training for attention but we learn to do so as we perform the task over a period of time. In our approach, first a driver is trained by supervised learning, to be able to predict the driving actions of steering angle, throttle and brake from the road scenes as input images. Then, unsupervised learning is used to train a saliency module which takes as input an image and produces a saliency incorporated image as output. The learning of this visual attention model is unsupervised in the sense that no eye gaze data is used for the purpose of training as illustrated in the Figure 1. This training framework has two goals to achieve; firstly, the driver model trained in the first step should be able to drive equally well on the saliency incorporated image as on the original image and secondly the number of pixels considered as salient by the attention model should be low. This is explained in greater detail in Section 3.

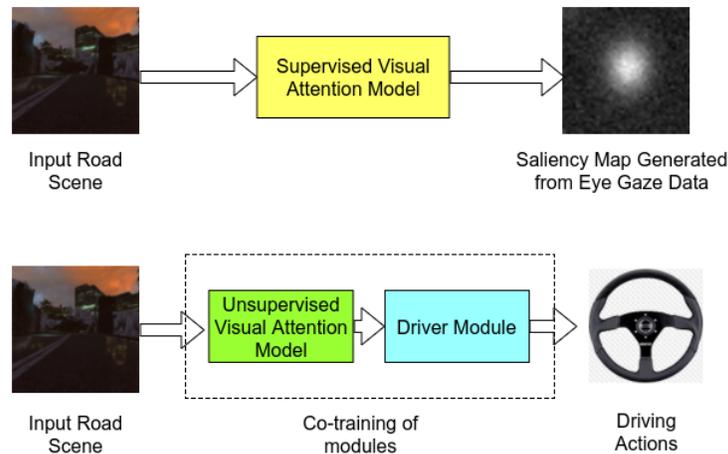

Figure 1. The two approaches for training the visual attention models; (top) supervised approach using eye gaze data and (bottom) unsupervised approach without involving any eye gaze data.

As next step to our method of solving the problem at hand we used a multiplication-based approach for incorporating the saliency in the final autonomous driving agent. In this approach, the original road scene images which are composed of three channels (RGB) was multiplied pixel-wise (for all the three channels) with its saliency map (single channel) generated by the attention models, and this saliency multiplied version of the road scenes was used for training the autonomous driving agent to predict the various driving actions. This multiplication approach is illustrated in Figure 2.

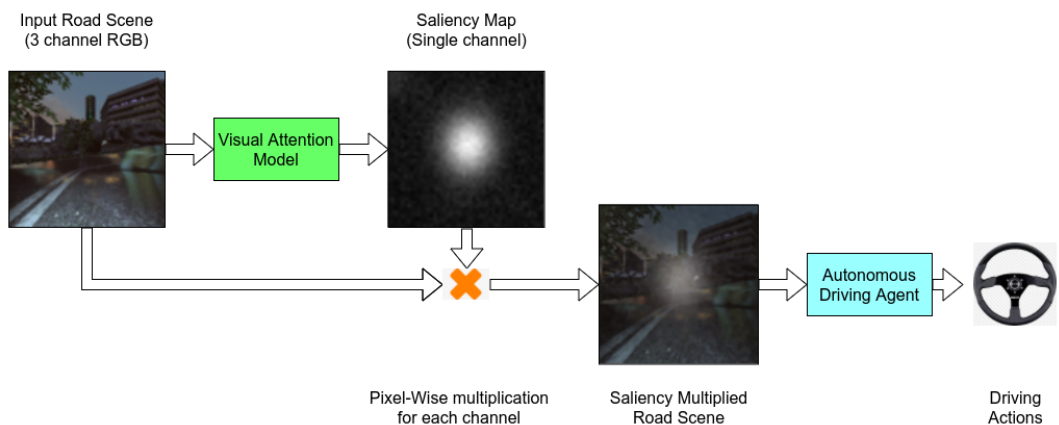

Figure 2. Illustration of multiplication approach for incorporation of visual attention in autonomous agent.

[1]Website : https://www.tobiipro.com/

In summary we introduce the novel technique of solving the autonomous driving problem with incorporation of saliency as an input to the agent. We finally present the results of our studies, which clearly shows that the proposed methodology performs better when evaluated in accordance with standard metrics. The supervised saliency model coupled with the multiplication method for incorporation performs the best among all the approaches.[2]

## 2. SUPERVISED LEARNING OF VISUAL ATTENTION FOR ROAD SCENES

In this section we describe the supervised approach used for training the visual attention model for road scenes while driving. As mentioned earlier the incorporation of visual attention in the autonomous driving agent is likely to help in reducing the scene intractability by predicting which parts of the road scene in front of the vehicle are more salient than the others. Since visual attention of road scenes in case of driving as a specific task is remarkably different from visual attention/saliency in natural images, we generated our own dataset to develop the visual attention model. The details of this data set are described below.

### 2.1 Eye Gaze Data Set used for Supervised Learning of Visual Saliency

The dataset used for training the supervised visual attention model was built for the specific task of driving. The data collection process involved collecting eye gaze data while the task of driving was performed simultaneously by the user in the simulator, Deepdrive. The Deepdrive simulator was used to record frames from the front facing hood camera of the car and the open source software OGAMA(OpenGazeAndMouseAnalyzer) for corresponding eye gaze data collection. In order to synchronize the two asynchronous means of data collection, we took advantage of the additional keyboard and mouse logs from the OGAMA(OpenGazeAndMouseAnalyzer) software. Thus, the raw data consisted of the x and y coordinates of the gaze points and image frames.

The frequency of data collection of the eye tracking setup comprising of Tobii Pro and OGAMA were much faster than the simulator, namely Deepdrive. So, we in order to find the gaze point for a particular frame, we stored the computer time stamp of the simulator recordings and followed the following approach thereafter. For each frame, we found the corresponding gaze point according to time stamps. And then apart from that gaze point, we also took the previous four gaze points. This step enabled us to the model the action specific visual attention in a better way as the action for a particular frame was guided by few of the previous frames. So, for each frame we had five gaze points. Then we used the following heuristic. For each of the five points, apart from the original one, which is the reference for the frame at hand; we found out the distance of that point from the reference point for that frame. If the point was within a certain threshold (of 10 pixels), we included that point in computation of the weighted mean, else it was rejected as being an out-lier. Finally, the weighted mean was computed with the reference point receiving the maximum weight and others a gradually decaying weight. Thus, we were left with one gaze point for each frame.

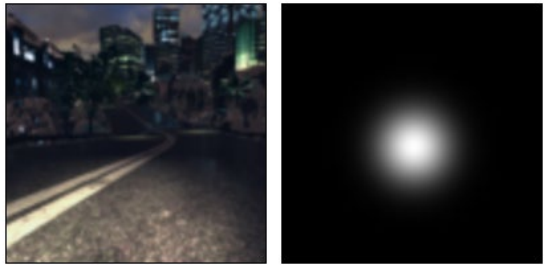

Figure 3. The figure shows an example of an image of road scenes and its corresponding saliency map in the dataset developed using eye gaze data.

The next step was to convert the gaze points to a saliency map for each frame. To this end we assumed a normal distribution around the final gaze point for each frame. The mean of the normal distribution was the final gaze point itself and the standard deviation was assumed to be a fixed value 20 pixels out of 227 pixels (this value of 227 pixels is specific to the image dimensions as returned by the simulator Deepdrive). Then the saliency of each pixel was computed using the following formula.

$$Saliency((x,y)) = e^{((-1) \times ((x-\mu_x)^2 + (y-\mu_y)^2)/2\sigma^2)} \qquad (1)$$

where,

$x$ = X coordinate of the pixel

$y$ = Y coordinate of the pixel

$\mu_x$ = X coordinate of the gaze point estimated for that frame

$\mu_y$ = Y coordinate of the gaze point estimated for that frame

$\sigma$ = Standard deviation of the above assumed Gaussian distribution

The final data set had a saliency map corresponding to each frame, Figure 3. Since drivers usually focus on the end of the road which lies close the center, a large number of images had their salient region close to the center. This "central bias" in the dataset was also observed in [15] and also in [16]. In order to attenuate the effect of the "central bias", we applied a transformation heuristic, which is described in the next subsection.

## 2.2 Translation Heuristic to Remove Central Bias

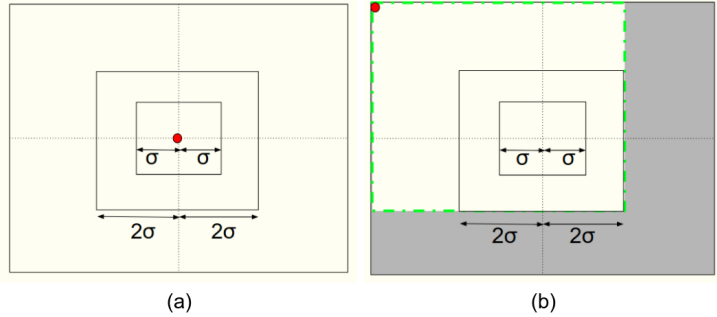

Figure 4. The above illustration is used to explain the translation heuristic. In both the diagrams the red dot depicts the center of the image coordinate axes, used in them. In part (a), a single frame of the road scene is shown, two squares are drawn centered around the coordinate axes, with dimensions of $2 \times \sigma$ and $4 \times \sigma$, where $\sigma$ represents the standard deviation assumed for the spread of visual attention as mentioned above. Part (b) shows the crop generated (using the first corner) in case a frame is found to be contributing to the central bias. The area bounded by the dotted green box depicts the cropped part of the frame that is added to the existing dataset after rescaling and the parts shaded with gray denote the part which is discarded. Similarly, the remaining three crops are generated using the other three corners.

For each frame in the data set that has been described above, we had the gaze point corresponding to that frame. We defined a square region about the center of the frame. The length of the region was equal to twice of the standard deviation, $\sigma$ used for computing the saliency map (i.e. $2 \times 20 = 40$) pixels out of the 227 pixels (this is specific to the dimension of the images returned by the Deepdrive simulator). In order to determine whether or not a particular frame contributed to the central bias, we computed whether the gaze point for that frame lay within the above defined square region. If it did, then we generated four crops from the four corners of the frame and added them to the existing data set after applying appropriate central coordinate and extrapolation measures. This process is illustrated in the Figure 4. On the other hand, if the gaze point lied outside the square region, no addition to the existing data set. This process not only enabled us to remove the central bias from the data set but also helped augment the existing data set with additional data points.

## 2.3 The Visual Attention Model for Road Scenes

This visual attention model considered modeling visual attention at a pixel level, such that each pixel in the frame is assigned a saliency value, which when normalized over all pixels in a frame for the road scene corresponds to the probability of a driver's gaze point falling on that pixel.

This model, (referred to as RoadSal henceforth) was inspired from the existing visual attention models, including SALGAN and SalNet. The architecture of this model as shown in Figure 5 comprised of 3 convolutional layers each of which was followed by a max pooling layer. We used a limited number of convolutional layers in order to avoid overfitting. Also, the original images were reduced to a much smaller size of [96 $\times$ 96], to avoid overfitting. The max pooling layers enabled us to reduce the size of feature maps further. The activation function used in each of the convolutional neural network was the Rectified Linear Unit (ReLU). The output from the third max pooling layer was flattened and then the max among pairs of them was taken. This helped prevent overfitting and also reduced the size of the network. The final layer was fully connected with linear activation function. The final output vector had a dimension of 2304, which when mapped into a 2D array of [48 × 48] corresponds to the visual attention map.

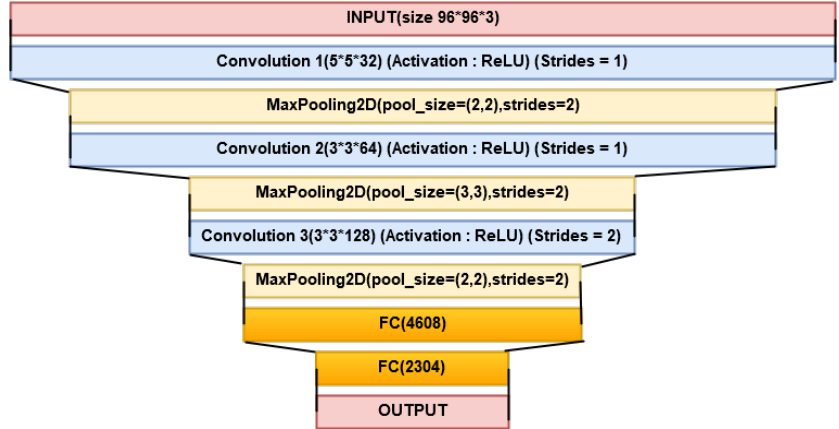

Figure 5. The neural network architecture for the supervised visual attention model, RoadSal.

The model, RoadSal was trained from scratch. The data set described above was directly used for the purpose, with the image frames as the input and the visual attention maps as the output. There were 28,000 data points in the entire data set. We used 80% of this dataset for training purposes. Stochastic gradient descent was used as the optimizer with a learning rate of $e^{-3}$, momentum of 0.9 and decay of 0.005. A batch size of 300 was used and cosine proximity was used to formulate the loss function. Since humans usually pay attention to small regions in the image, this leaves a significant portion of pixels being labeled as non-salient leading to a highly skewed dataset with heavy class imbalance. A mean squared error loss function is likely to perform poorly on this kind of data and hence we used a cosine loss function instead to mitigate the effect of this skewness. The loss function is described below:

$$Cosine\_Loss = -\frac{\overrightarrow{PSM} \cdot \overrightarrow{ASM}}{\left\|\overrightarrow{PSM}\right\| \times \left\|\overrightarrow{ASM}\right\|} \qquad (2)$$

where,

$\overrightarrow{PSM}$ = Flattened Predicted Saliency Map vector

$\overrightarrow{ASM}$ = Flattened Actual Saliency Map vector

### 3. AUTOTASKSAL: A FRAMEWORK FOR UNSUPERVISED LEARNING OF VISUAL ATTENTION PREDICTION MODEL

The collection of data for supervised task-specific attention prediction is cumbersome, costly and not completely accurate. Thus, an unsupervised model can be beneficial for task-specific attention modeling. We propose a training framework (referred to as "AutoTaskSal") for unsupervised learning of a visual attention model.

### 3.1 Description and Training

There are two neural networks used in this framework, AutoTaskSal. The first one is the saliency prediction model which (referred henceforth as Net1) is a fully convolutional neural network (Figure 6). The second one is the driver model which (referred henceforth as Net2) is a neural network (Figure 7) which takes as input a road scene and predicts driving actions. First, we train the driver model as is normally done for behavioral cloning (Figure 8) (Note that training of this driver model (Net2) requires a supervised approach). After the Step 1 is completed (Figure 8) the weights of the driver model (Net2) are fixed. Then we train the visual attention (or saliency) model, Net1 as illustrated in (Figure 9).

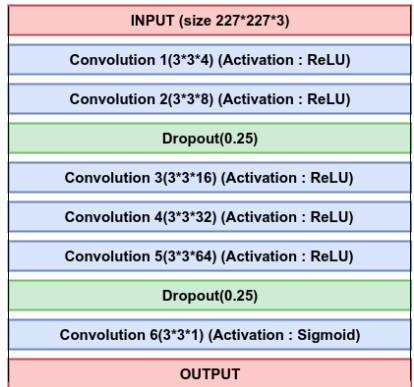

Figure 6. Net1: The neural network consisting only of convolutional layers, used to predict an attention map of the same height and width as that of the image.

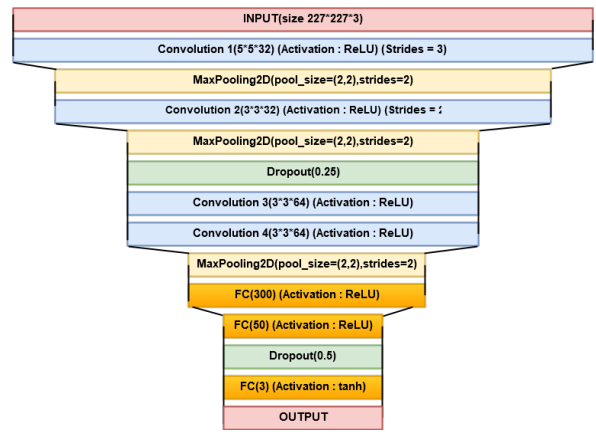

Figure 7. Net2: The neural network used to predict driving actions from the Road-Scene images.

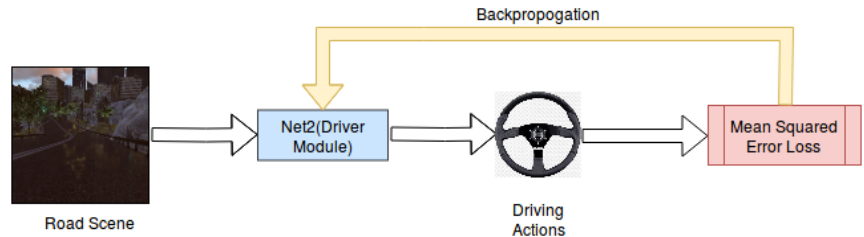

Figure 8. Step 1 of training in the AutoTaskSal framework where the driver model is trained with road scene images as input and driving actions as output. (Note, that this part is supervised behavioral cloning)

The predictions of the visual attention model help to enhance pixels in the input road scene in proportion to their importance in the driver's decision-making process. This training of Net1 (Visual Attention Model) is unsupervised in the sense that no labeled dataset (road scene and corresponding saliency map) is used for the purpose. The actual loss function is a linear combination of two loss functions. The first one (LOSS1), is to encourage the saliency model to discard pixels and the next one (LOSS2) is to make sure the driver still gets enough information to predict driving actuator values accurately. The saliency model first generates an attention map of the same size of the image. This attention map is then multiplied with the input road scene and given to the driver module which predicts the driving actions (actuator values, steering angle, throttle and brake) from this attention multiplied road scene.

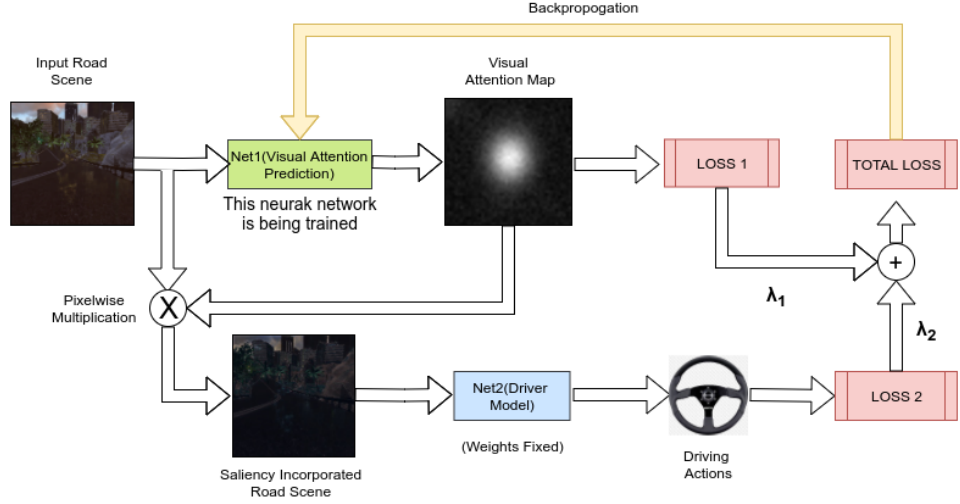

Figure 9. Step 2 of training in the AutoTaskSal framework for unsupervised training of the Visual Attention prediction model (Net1).

## 3.2 Loss Function

The first loss function 'LOSS1' tries to reduce the number of pixels being considered as visually salient. It is calculated as the average of square of all pixel-wise attention values. In each backpropagation step, this loss induces an update trying to reduce the attention awarded to each pixel by the saliency module. Its effect can be compared to the effect of regularization. It is calculated as follows:

$$LOSS1 = \frac{\sum_{i=1}^{height} \sum_{j=1}^{width} Attention[i][j]}{height \times width} \quad (3)$$

where,
height = height of attention map (here 227)
width = width of attention map (here 227)
Attention = 2D array of attention values generated by the attention module

The second loss function 'LOSS2' is simply the mean squared error of the predicted driving actions (steering, throttle and brake). This ensures that those pixels which are required for the driver module to predict driving actions accurately are retained. It is calculated as:

$$LOSS2 = \frac{(S_p - S_g)^2 + (T_p - T_g)^2 + (B_p - B_g)^2}{3} \quad (4)$$

where,
$S_p$ = Predicted Steering Value
$S_g$ = Ground Truth Steering Value
$T_p$ = Predicted Throttle Value
$T_g$ = Ground Truth Throttle Value
$B_p$ = Predicted Brake Value
$B_g$ = Ground Truth Brake Value

'TOTAL LOSS' is a linear combination of the above two losses:

$$TOTALLOSS = \lambda_1 LOSS1 + \lambda_2 LOSS2 \qquad (5)$$

where,

$\lambda_1$ = weight assigned to LOSS1

$\lambda_2$ = weight assigned to LOSS2

The idea behind such a loss function is that, by reducing the number of pixels and hence amount of information presented to the driver module (through LOSS1) while simultaneously minimizing the error in prediction of driving actions, the attention module (Net1) would learn to predict the regions of the image (pixels) which are essential for driving as salient.

## 4. INCORPORATION OF VISUAL ATTENTION IN BEHAVIORAL CLONING BASED AUTONOMOUS DRIVER

In this section we describe the approach that was used for the incorporation of visual attention in the autonomous driving agent. We adopted behavioral cloning approach for the autonomous driving model, with the following components:

- **Input**: Image of the Road Scene in front of the vehicle
- **Output**: The various driving actions comprising of the steering angle, throttle and brake in case of the simulator in our work namely Deepdrive

As described in the previous sections (2 and 3), we trained and developed two different visual attention models. First, is the supervised attention model, RoadSal. The second attention model is the Net1 from the unsupervised learning framework, AutoTaskSal.

Next, we train three different models (the actual autonomous driving agent) using the behavioral cloning framework, with exactly the same architecture and hyper-parameters but with different inputs, with each model trying to predict the same driving actions (steering angle, throttle and brake). The raw data comprising of road scene frames and driving action (steering angle, throttle and brake) pairs was collected using the oracle path follower of the simulator Deepdrive. Below, we describe the three autonomous agents and their training (Figure 10):

- **Model1**: This model (henceforth referred to as Model1) was trained with original road scene images as input to predict the driving actions (steering angle, throttle and brake) as output.
- **Model2**: The input for this model (henceforth referred to as Model2) involved the incorporation of visual attention predicted by RoadSal. We multiplied the saliency value pixel-wise for each of the three channels in the original road scene (image). This saliency multiplied image was the input for Model2 and the output were the driving actions (steering angle, throttle and brake).
- **Model3**: This model (henceforth referred to as Model3) also used saliency multiplied images as in case of Model2 for input and the driving actions (steering angle, throttle and brake) were the output. The difference from Model2 was that in this case the predictions of Net1 (component of AutoTaskSal) were used as the saliency maps.

## 5. RESULTS

We compared the performance of the autonomous driving agents Model1, Model2 and Model3 (Figure 10).

### 5.1 Simulator

For the purpose of studying the end results of the various models, we relied on Deepdrive as the simulator. The test data was collected using its oracle path follower, which has road scenes as input and driving actions as output. Since, the inputs for training of Model1, Model2 and Model3 were different, hence while testing the original road scenes were passed through appropriate pipelines before it was fed to the actual autonomous driving agent.

### 5.2 Evaluation Metric

We used Mean Squared Error (MSE) as the evaluation criteria. Mean Squared Error (MSE) was computed for the three driving actions of steering angle, throttle and brake separately. The final error value was the average of these three error values. The mean squared error is calculated as follows:

$$MSE = \frac{(S_p - S_g)^2 + (T_p - T_g)^2 + (B_p - B_g)^2}{3} \quad (6)$$

where,

$S_p$ = Predicted Steering Value

$S_g$ = Ground Truth Steering Value

$T_p$ = Predicted Throttle Value

$T_g$ = Ground Truth Throttle Value

$B_p$ = Predicted Brake Value

$B_g$ = Ground Truth Brake Value

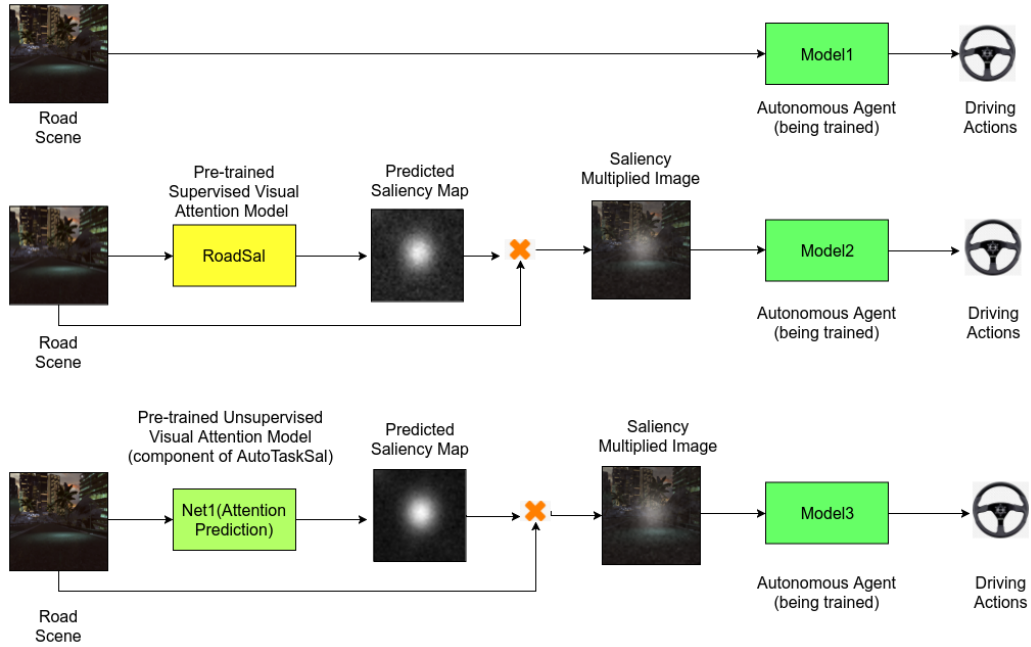

Figure 10. This figure describes the training for the three autonomous agents; Model1. Model2 and Model3.

### 5.3 Comparison

The mean squared error values over the same test set for the three models, Model1, Model2 and Model3 as illustrated in (Table 1), suggest that Model2 performs the best. The fact that Model2 performs better than Model1 clearly demonstrates the usefulness of incorporating of task specific visual attention in the context of autonomous driving.

| Autonomous Agent | Mean Squared Error (MSE) |
|---|---|
| Model1 | 0.01369 |
| Model2 | 0.01145 |
| Model3 | 0.034 |

Table 1: The different autonomous agents and their respective mean squared error values.

# 6. CONCLUSION AND FUTURE WORK

In conclusion, we developed a supervised visual attention model (RoadSal) for road scenes while driving using eye gaze tracking data. We also proposed an unsupervised learning framework for task specific visual attention (AutoTaskSal), and developed a visual attention model using it, which is Net1. Thereafter, we used a pixel-wise multiplication-based approach for incorporation of visual attention in the autonomous agents.

The incorporation of visual attention predicted by the supervised model, RoadSal led to the best performing autonomous driving agent. Thus, the incorporation of visual attention indeed improved the performance of the autonomous driver. This can be attributed to original motivation that human drivers pay different levels of attention to various things in front of them while driving. Also, it can be seen that using this approach we already inform the driver what is more important for decision making and hence better performance. Some examples of the visual attention incorporation are presented in (Figure 11).

As a future work, we would like to explore visual attention at an object level rather than at a pixel level as in the current work.

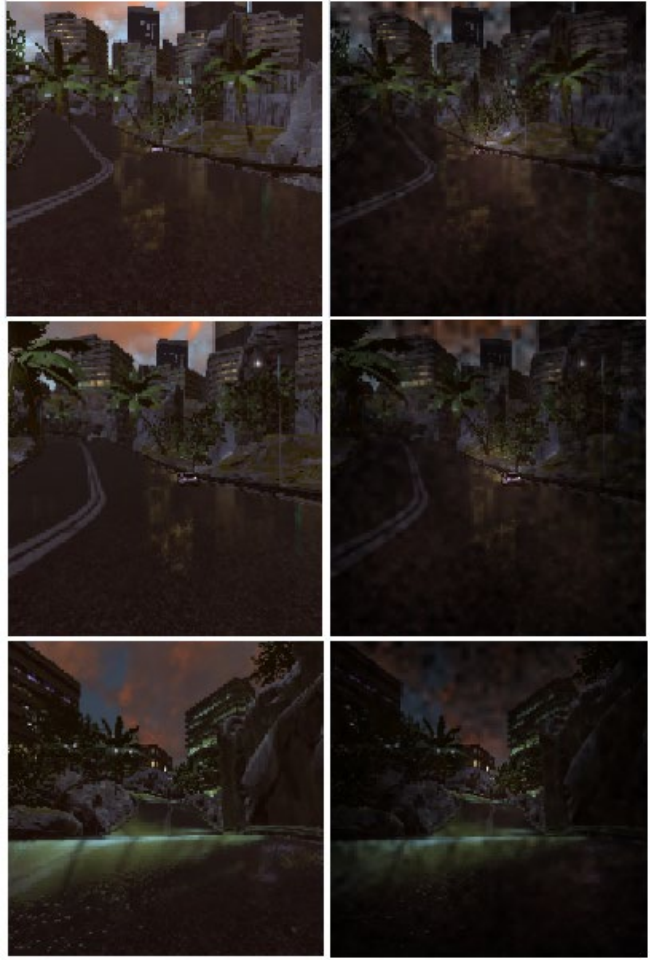

Figure 11. Examples of frames before multiplying saliency(left) and corresponding frames after multiplying predicted saliency from RoadSal(right). We can observe the region containing the car becoming more salient in the first two pictures. In the third picture, the buildings around the road receive less saliency as compared to the road.